# Remote Sensing of Forests using Discrete Return Airborne LiDAR

Hamid Hamraz[a] and Marco A. Contreras[b]

a: Department of Computer Science, b: Department of Forestry
University of Kentucky, Lexington, KY 40506, USA

**Abstract:** Airborne discrete return light detection and ranging (LiDAR) point clouds covering forested areas can be processed to segment individual trees and retrieve their morphological attributes. Segmenting individual trees in natural deciduous forests however remained a challenge because of the complex and multi-layered canopy. In this chapter, we present (i) a robust segmentation method that avoids a priori assumptions about the canopy structure, (ii) a vertical canopy stratification procedure that improves segmentation of understory trees, (iii) an occlusion model for estimating the point density of each canopy stratum, and (iv) a distributed computing approach for efficient processing at the forest level. When applied to the University of Kentucky Robinson Forest, the segmentation method detected about 90% of overstory and 47% of understory trees with over-segmentation rates of 14% and 2%. Stratifying the canopy improved the detection rate of understory trees to 68% at the cost of increasing their over-segmentations to 16%. According to our occlusion model, a point density of ~170 pt/m² is needed to segment understory trees as accurately as overstory trees. Lastly, using the distributed approach, we segmented about two million trees in the 7,440-ha forest in 2.5 hours using 192 processors, which is 167 times faster than using a single processor.

## 1. Introduction

Global forests cover about 30% of the land surface of the earth, include 80% of plant biomass, and account for 75% of primary productivity of biosphere, providing essential and unreplaceable ecosystem services to humans and the life on the planet [1]. Decision making in forest management has traditionally been based on stand (forested area with similar vegetation characteristics) attributes collected using field sampling and interpretation of aerial photography [2]. Field sampling is inevitably limited to a small percentage because of the large acquisition costs, which results in rough estimates of stand attributes while ignoring large variability in terrain and vegetation within stands [3]. Recent advances in remote sensing, geographic information systems, and information science have the potential to bring dramatic changes to forest data acquisition and management by providing inventory information at unprecedented spatial and temporal resolutions. Specifically, discrete return airborne light detection and ranging (LiDAR) technology has extensively been used in the past two decades in forestry [4, 5]. Due to its ability to penetrate vegetation canopy, LiDAR data captured in the shape of 3D point clouds contain vertical information from which vegetation structural characteristics can be retrieved, even from understory canopy layers. Airborne point clouds has successfully been used to derive digital surface models representing top surface of objects above the ground, as well as digital elevation models (DEMs) representing bare ground surface excluding the above ground objects [6]. Earlier studies built relationships between the LiDAR-derived variables and vegetation attributes using field sampling and extrapolated those relationships to the LiDAR coverage area and/or to the entire forest [7]. Although such methods can remarkably reduce field work and increase precision, they are insufficient for detailed forest management at the individual tree level [8].

Detailed tree-level forest management activities require individual trees to be segmented from the LiDAR point clouds. Although numerous tree segmentation methods have been developed, they have majorly focused on conifer forests or forests with relatively open canopy where assumptions about size and shape of tree crowns are made [9]. Deciduous forests present considerably more complex vegetation conditions due to large variation in tree shapes and sizes, larger number of species, and denser canopy where individual trees are considerably more challenging to segment [8]. In addition, retrieval of understory trees using airborne LiDAR is much harder because of the reduced amount of LiDAR points penetrating below the main cohort formed by overstory trees [10]. Although understory trees provide limited financial value and a minor proportion of total above ground biomass, they influence canopy succession and stand development, form a heterogeneous and dynamic habitat for numerous wildlife species, and are an essential component of ecosystem functioning [11]. Typically, detection rate of overstory (dominant and co-dominant) trees is above 90% and the detection rate of understory (intermediate and overtopped) trees is below 50%. Although variability in stand structure and terrain condition is the major factor affecting tree segmentation quality [9, 12], a minimum point density is the basic requirement for a reasonable segmentation of trees [13, 14]. This basic requirement is typically not satisfied for understory trees in a dense forest. Furthermore, LiDAR data covering an entire forest is much more voluminous than the memory of a workstation and may also take an unreasonable time to be sequentially processed using an external memory algorithm. Because large-scale LiDAR data is typically arranged in several tiles for efficient management and delivery, distributed processing of different tiles seems to be straightforward. However, the data representing tree crowns located across tile boundaries are split into two or more pieces that need to be processed by different computing units. Only few studies have considered distributed processing of large geospatial data addressing the boundary problem [15] – specifically there are no studies considering forest data. This is increasingly important when obtaining tree-level information for areas other than small plots, which is often the ideal objective. Moreover, continuous advancements of sensor technology and platforms [16] is resulting in point clouds to be acquired with greater resolutions, increasing the need for more efficient and scalable processing schemes.

To address these limitations, in this chapter we present: i) a robust tree segmentation method, which is designed to be applied in multi-storied, closed-canopy deciduous forests, ii) a vertical canopy stratification procedure that separates the canopy into an overstory layer and multiple understory canopy layers, iii) a LiDAR pulse occlusion model in terms of point density to further investigate the inferior segmentation quality of understory trees, and iv) a distributed computing approach that addresses the tile boundary problem to efficiently and accurately process the data at the forest level.

## 2. Methodology

## 2.1. Robust tree segmentation

The main inputs of the tree segmentation method [17] are the LiDAR point cloud and the LiDAR-derived DEM. A pre-processing step is firstly applied to select the points representing the top of the surface, hereafter called LiDAR surface points (LSPs). The pre-processing step bins the point cloud to a horizontal grid with cell width equals to the average footprint (AFP) of the point cloud (equals to the reciprocal of square root of the point density) and selects the highest point within each grid cell as the LSP. Using the DEM, heights above ground are calculated for all LSPs and ground points in the LSPs are removed from further analysis. Based on the vegetation structure, this creates some gaps with no vegetation in the remaining LSP dataset, which are utilized later for segmentation. The last pre-processing step smooths LSPs to reduce small variation in vegetation elevation within tree crowns while maintaining important vegetation patterns. A Gaussian smoothing filter with standard deviation of 2×AFP was used.

After the pre-processing steps, the tree segmentation method consists of the following routines: 1) locate the global maximum (GMX) amongst LSPs, which is assumed to represent the apex of the tallest tree within a given area, 2) generate vertical profiles originating from the GMX location and expanding outwards, 3) identify the individual LSP along each profile that represents the crown boundary using between-tree gap detection and local minimum (LM) inspection, 4) create a convex hull of boundary points delineating the tree crown, and 5) cluster all LSPs encompassed within the convex hull and assign them as part of the current tallest tree crown. These routines are applied iteratively until all LSPs have been clustered into tree crowns. Clusters representing crowns with diameter below a minimum detectable crown width (MDCW), set here as 1.5 m, are considered noise. Also, clusters entirely below a minimum height of 4 m, are also removed as they are likely associated with ground-level vegetation. Figure 1-a shows the flowchart of the tree segmentation method and Figure 1-b illustrates an iteration of a tree clustering.

### 2.1.1. Profile generation

After identifying the GMX, eight uniformly spaced vertical profiles (every 45°) originating from it and expanding a maximum horizontal distance of 20 m are generated. The width of each profile was set to 2×AFP to ensure a sufficient number of LSPs representing vegetation characteristics. After the crown boundary is determined for each profile (explained below), the maximum crown radius ($r$) is used to determine the chord height ($x$) between two maximum crown radii profiles separated by the angular spacing ($\varphi$) (Figure 2) using the following equation:

$$x = r(1 - cos(\varphi/2)) \qquad (1)$$

If the chord height is larger than the AFP, the angular spacing is reduced to half and the number of profiles is doubled. The new chord height is calculated again based on the updated maximum crown radius and the new profile angular spacing. Doubling the number of profiles iterates until the chord height becomes smaller than the AFP. In this way, the number of profiles adapts incrementally according to the maximum LiDAR-derived crown radius to smoothly delineate the tree crowns.

### 2.1.2. Crown boundary identification

To identify the crown boundary along a profile, inter-tree crown gaps are firstly detected. These gaps are assumed to correspond to relatively large horizontal distance between two consecutive LSPs along the profile, which are identified using a statistical outlier detection method. Horizontal distances between consecutive points along the profile follow a Poisson distribution. When transformed to their square root or to their logarithm, the resulting distribution reasonably approximates a normal distribution, which is more appropriate for the outlier detection method. To be conservative, transformed distance values larger than six times the interquartile range from the third quartile are considered inter-tree gaps. After gaps are detected, only the LSPs between the GMX and the first gap remain for further analysis and the LSPs located beyond the first gap are removed from the profile. At this point, the profile is assumed to represent either the crown of the current tallest tree or the crowns of multiple adjacent trees, growing close together with overlapping crowns.

In the next step, the remaining LSPs are inspected starting from the GMX to locate an LM, which is defined as an LSP with elevation lower than its two adjacent LSP neighbors. Once an LM is found, two windows expanding on both sides of it are created to determine if the LM represents the crown boundary. The left window considers all LSPs from the GMX to the LM. The size of the right window is estimated based on the steepness of the crown ($S_{right}$), which is calculated as the median (in degrees) of the absolute slopes between consecutive points ($i, i+1$) located within a distance of MDCW to the right of the LM ($MDCW_{right}$):

$$S_{right} = tan^{-1}\left(median\left[\,|slope_{i,i+1}|\,|\,i,i+1 \in MDCW_{right}\,\right]\right) \qquad (2)$$

If the LM represents the crown boundary, the LSPs within $MDCW_{right}$ partially represent the crown of an overlapping and shorter tree with a steepness approximated by $S_{right}$. We bound the value of $S_{right}$ between the steepness of two distinct crown shapes: a sphere-shaped crown and a narrow cone-shaped crown (two ends of the spectrum). As the height of the adjacent tree ($h_{ad}$) is between the heights of the GMX and the LM point, it is approximated by the average of the GMX and the LM heights. The steepness of a narrow cone-shaped crown can be expressed as 90°-$\varepsilon$, where $\varepsilon$ (set here as 5°) indicates a small deviation from nadir. The cone-shaped crown radius ($cr_c$) can then be calculated as follows:

$$cr_c = \frac{h_{ad} \times CL_c}{tan(90° - \varepsilon)} \times O_c \qquad (3)$$

where $CL_c$ is the ratio of the crown height to the total height, and $O_c$ indicates the crown radius reduction due to the overlap assuming the cone-shaped tree is situated in a dense stand. On the other hand, the slope of a sphere-shaped crown ranges from 0° to 90° with the expected value of 32.7°. Its crown radius ($cr_s$) can be calculated as follows:

$$cr_s = \frac{h_{ad} \times CL_s}{2} \times O_s \qquad (4)$$

where $CL_s$ and $O_s$ indicate the crown to total height ratio and the crown radius reduction due to the overlap within a dense stand for the sphere-shaped tree.

The size of the right window ($w_{right}$) is then calculated by interpolating $cr_c$ and $cr_s$ with respect to $S_{right}$, which is bounded between 32.7° and 90°-$\varepsilon$, as follows:

$$w_{right} = cr_c \left( \frac{(90° - \varepsilon) - S_{right}}{(90° - \varepsilon) - 32.7°} \right) + cr_s \left( 1 - \frac{(90° - \varepsilon) - S_{right}}{(90° - \varepsilon) - 32.7°} \right) \quad (5)$$

Lastly, after determining both window sizes, the median of slopes between consecutive LSPs of each window is calculated. If the median slope of the left-side window is negative (downwards from the apex to the crown boundary) and the median slope of the right-side window is positive (upwards from the crown boundary toward the apex of the adjacent tree crown), the LM is considered a boundary point. Otherwise, the current LM is considered to represent natural variation of vegetation height within the current tallest tree crown and the next LM farther from the GMX along the profile is evaluated. If none of the LMs found meets the crown boundary criterion, the last LSP is considered as the crown boundary.

The crown height to total height ratio is highly variable among individual trees and species with values typically varying between 0.4 and 0.8 [18]. The crown ratio of a narrow cone-shaped tree tends to be larger than that of a sphere-shaped one. Thus, for the purpose of illustrating the application of our method, we used 0.8 and 0.7 for CLc and CLs, respectively. Similarly, crown radius reduction due to overlap is highly variable with a value of even less than one half for a very dense stand. The radius of a narrow cone-shaped tree tends to be reduced less than of a sphere-shaped tree because the crown of a narrow cone-shaped tree is quite compact from the sides. Thus, we used two thirds for Oc and one third for Os. . Although these values can affect the size of the right window (Equation (5)), the sign of the median slope would remain the same as long as the size is within a reasonable range. Also, considering the multiple profiles generated for each GMX, the effect of a single potentially mislocated boundary point on the ability to delineate tree crown is reduced.

Both routines for generation of a sufficient number of profiles and for identifying the crown boundary of each profile are completely based on the information extracted on-the-fly from the 3D locations of the LSPs. This avoids a priori assumptions of tree crown shapes, dimensions, and inter-crown spacing, which yields a robust tree segmentation method that can be applied to different vegetation types.

## 2.2. Vertical canopy stratification

The canopy stratification procedure [19] uses all LiDAR points binned into the horizontal grid. It then analyzes the height histogram of all LiDAR points within a circular locale around each individual grid cell. The locale should include a sufficient number of points for building an empirical multi-modal distribution but not extend too far to preserve locality. We fixed the locale radius to 6×AFP (essentially containing about $\pi \times 6^2$ points), which is lower bounded at 1.5 m to prohibit too small locales capturing insufficient spatial structure. To analyze the height histogram of each locale, the histogram is smoothed using a Gaussian kernel with standard deviation of 5 m to remove variability pertaining to vertical structure of a single crown. The salient curves in the smoothed histogram (height ranges throughout which the second derivative of the smoothed histogram are negative) represent the canopy layers [20]. The midpoint of the top canopy layer and the canopy layer below it are regarded as the height threshold for stratifying the top layer in that cell location (Figure 3). Using the height thresholds determined for all grid cells, the top canopy layer is stratified and removed from the point cloud and the AFP is updated according to the density of the remainder of the point cloud. The stratification procedure iterates binning the remainder of the point cloud into a horizontal grid with the cell size equal to the updated AFP, analysing locales of individual grid cells, and removing layers until the point cloud is emptied.

As the height thresholds to stratify the top canopy layer in each iteration of the procedure are determined using overlapping locales, the canopy layer smoothly adjusts to incorporate the vertical variability vegetation height to minimize under/over-segmenting tree crowns across layers. To improve segmentation of understory trees within a point cloud representing a multi-story stand, the method presented in the previous section can be applied independently to each canopy layer (Figure 4).

### 2.3. Canopy occlusion model

Using the canopy stratification procedure, we model the occlusion effect of higher canopy layers on the lower layers in terms of point density [21], which is defined as the number of points divided by the horizontal area covered by them. Point density of the entire LiDAR point cloud (PCD) is dependent upon different flight and sensor parameters such as flight altitude and speed, pulse repetition rate, field of view, and swath overlap [14, 22]. These parameters also affect the fractions of points recorded for over/understory canopy layers, yet point density of individual layers decreases with proximity to ground level.

Assuming all canopy layers cover the same area as the entire point cloud, PCD equals the total of point densities of constituting canopy layers of the point cloud plus the density of the DEM. Because the ground is different from a canopy layer in interaction with LiDAR pulses, necessitating a different density model for the DEM, we assume an infinite number of canopy layers were placed instead of the ground to simplify the analysis. The point density of DEM approximately equals the total of point densities of the canopy layers in place of the ground. Hence PCD can be calculated as the sum of point densities of an infinite number of canopy layers (the actual ones plus those in place of the ground):

$$PCD = d_1 + d_2 + d_3 + \ldots + d_n \quad n \in \mathbb{N} \tag{6}$$

where $d_n$ denotes the point density of the $n$th top canopy layer and converges to zero as $n$ increases. To normalize point densities, we divide both sides of Equation (6) by PCD:

$$1 = p_1 + p_2 + p_3 + \ldots + p_n \quad n \in \mathbb{N} \tag{7}$$

where $p_n$ denotes the fraction of LiDAR points at the $n$th top layer and can be estimated using a logarithmic series distribution that has a discrete-domain decreasing function supporting natural numbers [23].

We denote the required PCD of a point cloud for reasonable segmentation of trees forming the top canopy layer of the point cloud by $PCD_{min}$. The required PCD of a point cloud for a reasonable segmentation of trees forming the $n^{th}$ top canopy layer can then be calculated using Equation (7). We hypothetically remove the $n-1$ top canopy layers of the point cloud. The resulting point cloud would have a density fraction of $1 - (p_1+p_2+\ldots+p_{n-1})$ of the original point cloud. Assuming this density fraction yields a density of $PCD_{min}$ for the resulting point cloud, the point density of the original point cloud for reasonable segmentation of trees forming its $n^{th}$ top canopy layer ($pcdmin(n)$) by proportionality becomes:

$$pcd_{min}(n) = \frac{PCD_{min}}{1-(p_1 + p_2 + \ldots + p_{n-1})} \tag{8}$$

## 2.4. Distributed processing of large-scale forest LiDAR data

The distributed approach for tree segmentation [24] uses a master-slave scheme, where the master is in charge of maintaining the global tile map and coordinating how to process individual tiles and their boundary data while the slaves perform the actual tree segmentation. Tile boundary data (solid/striped colored regions in Figure 5) represent tree crowns located between two tiles (light-colored) – hereafter referred to as edge data – or among three or four tiles (dark-colored) – hereafter referred to as corner data. After segmenting a tile, all segmented crowns that are within a horizontal distance of 2×AFP from a tile edge form part of the boundary data. Crowns adjacent to only one edge (solid light colored) are regarded as part of the associated edge data and those adjacent to exactly two edges (solid dark colored) are regarded as part of the associated corner data.

Figure 6 shows the flowcharts of the master and the slave processes, where it is assumed that all processes can independently input tiles data and output results. Such an assumption can reasonably be fulfilled by using a supercomputing infrastructure with a unified file system, or by maintaining the tiles and the results on a specialized distributed spatial data organization/retrieval system [25, 26]. The master initializes the work by loading the tile map and assigning each slave to process a unique tile via a process tile (PT) message carrying the associated tile ID. Upon receiving a PT message, a slave loads and segments the tile and identifies the boundary data inside the tile consisting of eight disjoint sets (four edges and four corners). The slave outputs the segmented non-boundary trees, notifies the master via a tile complete (TC) message carrying the boundary sets, and waits for the master for a new assignment. The master then updates the tile map and inspects all eight boundary sets it received from the slave to determine if any of the associated edge/corner data are ready to be unified. Edge data are ready when both tiles sharing the edge are segmented and corner data are ready when all four tiles sharing the corner are segmented. The master then unifies all edge/corner data that are ready and re-assigns the waiting slave to re-segment the unified boundary data, which is conveyed by a process boundary (PB) message to the slave. The slave process, upon receiving the PB message, segments the boundary data conveyed by the message, outputs the result trees, and notifies the master via a boundary complete (BC) message. The master then re-assigns a new tile via a PT message to the slave. If the master cannot locate any ready boundary data when it receives the TC message, it proceeds with re-assigning the waiting slave to segment a new tile via a PT message. If all tiles are segmented, the master terminates the slave process by sending a finalize (FIN) message. The master process continues until all slaves are finalized, implying that all tiles and their boundary data were processed.

In the presented distributed approach, all tile boundaries are guaranteed to be processed; once all tiles sharing each specific edge or corner are segmented, the edge/corner data are assigned to be processed by the slave that completed the last tile. Assuming that the amount of processing incurred by the master does not affect its responsiveness, the slaves keep working all the time resulting in an efficient distributed processing scheme (see Reference [24] for a theoretical analysis of the runtime and scalability).

## 2.5. Study site and LiDAR campaign

The study site is the University of Kentucky's Robinson Forest (RF, Lat. 37.4611, Long. -83.1555) located in the rugged eastern section of the Cumberland Plateau region of southeastern Kentucky in Breathitt, Perry, and Knott counties (Figure 7-a). RF features a variable dissected topography with moderately steep slopes ranging from 10 to over 100% facing predominately northwest to southeast, and with elevations ranging from 252 to 503 meters above sea level [27]. Vegetation is composed of a diverse contiguous mixed mesophytic forest made up of approximately 80 tree species with northern red oak (Quercus rubra), white oak (Quercus alba), yellow-poplar (Liriodendron tulipifera), American beech (Fagus grandifolia), eastern hemlock (Tsuga canadensis) and sugar maple (Acer saccharum) as overstory species. Understory species include eastern redbud (Cercis canadensis), flowering dogwood (Cornus florida), spicebush (Lindera benzoin), pawpaw (Asimina triloba), umbrella magnolia (Magnolia tripetala), and bigleaf magnolia (Magnolia macrophylla) [27, 28]. Average canopy cover across RF is about 93% with small opening scattered throughout. Most areas exceed 97% canopy cover and recently harvested areas have an average cover as low as 63%. After being extensively logged in the 1920's, RF is considered second growth forest ranging from 80-100 years old, and is now protected from commercial logging and mining activities.

The LiDAR acquisition campaign over RF was performed in the summer of 2013 during leaf-on season (May 28-30) using a Leica ALS60 sensor, which was set at 40⁰ field of view and 200 KHz pulse repetition rate. The sensor was flown at the average altitude of 214 m above ground at the speed of 105 knots with 50% swath overlap. Up to 4 returns were captured per pulse. Using the 95% middle portion of each swath, the resulting LiDAR dataset given the swath overlap has an average density of 50 pt/m². The provider processed the raw LiDAR dataset using the TerraScan software to classify LiDAR points into ground and non-ground points. The ground points were then used to create a 1-meter resolution DEM using the natural neighbour as the fill void method and the average as the interpolation method. The LiDAR dataset was delivered in 801 square (304.8 m side ~ 9.3 ha area) tiles (Figure 7-b), each containing about 5 million LiDAR points on average and occupying about 400 MB of disk space.

## 2.6. Evaluation

### 2.6.1. Field data

Throughout the entire RF, 270 regularly distributed circular plots of 0.04 ha in size spaced every 384 m were surveyed during the summer of 2013. Plot centers were georeferenced with 5 m accuracy. Within each plot, DBH (cm), tree height (m), species, crown class (dominant, co-dominant, intermediate, overtopped), tree status (live, dead), and stem class (single, multiple) were recorded for all trees with DBH > 12.5 cm. In addition, horizontal distance and azimuth from plot center to the face of each tree at breast height were collected to create a stem map. Site variables including slope, aspect, and slope position were also recorded for each plot. Table 1 shows a summary of the plot level data. LiDAR data corresponding to each plot include a 4.7-m buffer to capture complete crowns of border trees.

### 2.6.2. Tree segmentation evaluation

To evaluate the performance of the tree segmentation method, a score to each pair of LiDAR-derived tree location, assumed to be the apex of the segmented crown, and stem location measured in the field is assigned. The score is based on the tree height difference, which should be less than 30%, and the leaning angle between the crown apex and the stem location, which should also be less than 15° from nadir. The set of pairs with the maximum total score where each crown or stem location appears not more than once is selected using the Hungarian assignment algorithm and is regarded as the matched trees [17]. The number of matched trees (MT) is an indication of the tree segmentation quality. The number of unmatched stem map locations (omission errors – OE) and unmatched LiDAR-derived crown apexes that are not in the plot buffer area (commission errors – CE) indicate under- and over-segmentation, respectively. The segmentation accuracy is calculated in terms of recall (Re – measure of tree detection rate), precision (Pr – measure of correctness of detected trees), and F-score (F – combined measure) using the following equations [29]:

$$Re = \frac{MT}{MT + OE} \quad (9)$$

$$Pr = \frac{MT}{MT + CE} \quad (10)$$

$$F = \frac{2 \times Re \times Pr}{Re + Pr} \quad (11)$$

## 3. Results and discussion

### 3.1. Tree segmentation

On average for the 270 plots in RF, recall of the tree segmentation method was 60%, precision was 91%, and F-score was 70% (Figure 8). Recall values ranged from 16% to 100% and precision values ranged from 36% to 100%. In dense plots with a relatively large number of understory trees, several trees were missed resulting in relatively low recall values. However, more than 90% of overstory trees were detected in those plots. As expected, the overall three segmentation accuracy metrics were higher for overstory trees compared with understory trees (Figure 8). Recall, precision, and F-score for overstory trees were 91%, 85%, and 86%, respectively, while for understory trees they were 48%, 97%, and 62%. When considering all crown classes, the tree segmentation method was able to detect 95% of dominant, 90% of co-dominant, 62% of intermediate, and 30% of overtopped trees in the 270 plots. In addition, the method was able to detect 20% of dead trees.

By stratifying the canopy to layers and independently segmenting each layer, the average precisions over the 270 plots decreased for all crown classes while average recalls increased (Figure 8). When comparing performance measures with and without stratification using two-tailed paired T-tests, all metrics except F-score for overstory trees showed significant ($P < .0001$) changes, which is also evidenced by the non-overlapping 95% confidence intervals in Figure 8. Recall and precision for understory trees showed the most remarkable changes: an increase from 46% to 68% (MSE = 10.04) and a decrease from 99% to 84% (MSE = 3.97), respectively. Overall, vertical stratification of canopy improved the F-scores for understory (from 61% to 73%, MSE = 1.70) as well as for all trees (from 70% to 77%, MSE = 0.66), while it barely affected the F-score for overstory trees. When considering all crown classes, detection rate for dominant, co-dominant, intermediate, overtopped, and dead trees was 96%, 93%, 72%, 53%, and 30%, respectively. The improvements gained as a result of canopy stratification are mainly due to a strong increase in detection rate and a moderate decrease in correctness of the detected understory trees. This observation indicates an increased sensitivity to segment understory trees while not affecting the segmentation of overstory trees.

### 3.2. Stratified canopy layers

For most of the 270 plots, the canopy stratification procedure identified three (68.2%) or four (24.1%) canopy layers with an expected number of canopy layers of 3.16 per plot. Any layer entirely located below 4 m was excluded because it likely represents ground level vegetation, although any of the remaining layers may extend below 4 m and even touch the ground. Starting height and thickness of a canopy layer are defined as the medians over all grid cells used to stratify the layer (Figure 3). The average starting height of a canopy layer ranged from 0.3 to 18.2 m and the average thickness of a layer ranged between 6.1 and 8.8 m. Also, the average point density ranged between 0.44 and 42.08 pt/m² (Table 2). The average starting height, thickness, and point density of the entire canopy (all layers aggregated) was 1.4 m, 24.8 m, and 47.5 pt/m², respectively. The average point density of the entire canopy agrees with the average point density of the initial LiDAR dataset of 50 pt/m² given that ground and ground level vegetation returns were removed.

Thickness and point density generally decrease with lower canopy layers (Table 2). Specifically, the third and fourth top canopy layers, where the majority of understory trees are found, have an average density lower than 1 pt/m². Such low density is far less than the reported optimal point density in the literature of ~4 pt/m², where accuracies for segmenting individual trees plateau [13, 14]. This low density of understory canopy layers is the main reason for inferior tree segmentation accuracy of understory trees compared with overstory trees. As reported by Kükenbrink et al. [10], at least 25% of canopy volume remain uncovered even in small-footprint airborne LiDAR acquisition campaigns, which concurs with suboptimal point density of lower canopy layers for tree segmentation in our study.

### 3.3. LiDAR occlusion model

In order to estimate $p_n$ in Equation (7), we used the canopy stratification result of the 270 plot point clouds (Table 2). We recorded a sequence of five $p_n$ values ($1 \leq n \leq 5$, zeros for missing layers) for each point cloud. We then fitted a logarithmic series distribution to all $(n, p_n)$ pairs (N = 852, MSE = 0.0025 – Figure 9):

$$p_n = \frac{0.266^n}{-ln(1-0.266) \times n} \qquad n \in \mathbb{N} \qquad (12)$$

As mentioned, the reported optimal PCD for a reasonable segmentation of trees is about 4 pt/m² [13, 14], which is the value we adopt for $PCD_{min}$. Using Equation (8) where $p_n$ is calculated from Equation (12), the minimum PCD required to reasonably segment trees in as deep as the third canopy layers would be 169.57 pt/m². Similarly, if we require having a reasonable segmentation for as deep as only two canopy layers, the minimum PCD becomes 30.1 pt/m².

### 3.4. Distributed segmentation of entire forest

We implemented the distributed segmentation approach without canopy stratification using the message passing interface (MPI) and ran it on the University of Kentucky Lipscomb cluster. Using 192 processing cores, the distributed approach segmented a grand total of 1,994,970 trees in 2.5 hours. This time is actually ~167 times shorter than using only a single computer assuming the computer has enough memory to load the entire LiDAR dataset. Although in theory tile size does not affect the segmentation result of the distributed approach, a slight bias that was linearly correlated with the total length of the shared tile edges was introduced in practice. To quantify the bias, we created five 1.5×1.5 Km square blocks of LiDAR point clouds (including ~3% of the entire RF data per block) and segmented each block using the tree segmentation method run on a single computer. We then partitioned each block into grids of 2×2, 3×3, ..., 15×15 sub-blocks and segmented the partitioned blocks using the distributed approach. By comparing the number of trees segmented for different partitioning patterns, we discovered that an average of 96 additional trees (false positives) were segmented per 1 Km of shared sub-block edge length (see Reference [24]). Given the total length of the shared tile edges of the entire RF data is 446.23 Km (Figure 7-b), the estimated number of false positives became 42,833 (2.15% of the grand total number of trees). Subtracting these false positives resulted in a grand total of 1,952,137 segmented trees.

Due to the imperfectness of the segmentation method within each tile, a portion of the grand total number of segmented trees was over-segmented and a portion of existing trees in the forest was undetected. To account for the over-segmentations/undetected trees, we used the accuracy results of the segmentation method on the 270 plots of RF (Figure 8). The detailed accuracy result included the number of detected trees (bearing over-segmentations) and the number of existing trees (bearing undetected trees) by crown classes (dominant, co-dominant, intermediate, and overtopped). Within each of the 270 plots, we calculated by crown class a fraction of the existing trees in that crown class divided by the grand total (all crown classes) of detected trees. Table 3 shows the mean and 95% confidence interval bounds of the fractions across the 270 plots. It also shows the adjusted estimates of number of existing trees, which were calculated by multiplying the grand total number of detected trees using the distributed approach to the corresponding fractions. Considering a 95% confidence interval, the total number of existing trees in the 7,441.5-ha forested area is estimated to be 2,495,170 (±5.71%), which results in an average of 335.30 trees per ha.

For verification purposes, we compared the LiDAR-derived tree number estimates (Table 3) with equivalent estimates based on field measurements of the 270 plots from RF (Figure 10). The estimates for total number of trees per ha differ by about 3% (~342 LiDAR-derived compared with ~326 field estimated) and the estimates of number of dominant trees per ha differ by about 30% (~21 LiDAR-derived compared with ~15 field estimated). However, the large overlaps between the 95% confidence interval errors indicate no statistically significant differences. The height distribution of all segmented trees in RF shows a bimodal pattern (Figure 11), which can be attributed to the multi-story structure. A normal mixture model to the bimodal distribution was fitted, where the larger peak (associated with overstory trees) has a mean height of 26.9 m and a standard deviation of 6.6 m, and the smaller peak (associated with understory trees) has a mean height of 9.4 m and a standard deviation of 2.6 m. We also compared the LiDAR-derived mean tree height estimates with those obtained from the 270 sample plots (371 overstory and 826 understory trees, Table 1). The sample mean height of overstory trees was 25.4 m with a standard deviation of 5.3 m, and the sample mean height of understory trees was 17.0 m with a standard deviation of 4.1 m. Considering that the LiDAR-detected tree heights are in fact biased by presence of falsely detected trees and absence of undetected trees, the field estimates are close to the LiDAR-detected estimates for overstory trees. However, field estimates for understory trees are remarkably larger than the LiDAR-detected estimates. The reason is likely that the only information used to fit the normal mixture model was the heights of the trees while height may not be sufficient for classification, i.e., a moderately tall tree can be classified as an understory tree if it is surrounded by taller trees while the mixture model would always classifies it as overstory according to its height, and vice versa.

As the distributed processing approach uses the tree segmentation method as a building block and does not require any knowledge on how the method functions, it may be used to straightforwardly adopt any other single-processor object identification/segmentation algorithm to scale up processing arbitrarily big spatial and geospatial datasets, such as remotely sensed buildings, cars, planets, etc. The only caveat is that the objects may not be greater than the tile size (they may not touch more than two adjacent edges of a tile). Also, generalization of the approach to process 3D spatial data in a similar manner is possible. Instead of tiles that are representing surfaces, cubes representing volumes form the data units for 3D analysis. Boundary data in this case would be surface (shared between two cubes), edge (shared among four cubes), and corner (shared among eight cubes) that can be handled for distributed processing using the master-slave processing scheme.

## 4. Conclusions

Airborne LiDAR point clouds covering forested areas contain a wealth of information about the horizontal and vertical structure of vegetation. This information can be used to segment individual trees and subsequently retrieve their morphological attributes. However, existing tree segmentation methods are forest type specific and are majorly focused on conifer stands and/or forests with open canopy. We presented a robust tree segmentation method that does not require prior assumptions about the crown shapes/sizes nor tree spacing. The proposed method was defined as a generalizable approach that can be applied in deciduous natural forest with complex vegetation and stand structures. Results showed an overall segmentation accuracy of ~87% for overstory trees and ~61% for understory trees. We also presented a canopy stratification procedure to separate the point cloud to an overstory and multiple understory canopy layers. We applied the tree segmentation method independently to each canopy layer, which improved the overall segmentation accuracy of understory trees from ~61% to ~73% while barely affecting the segmentation accuracy of overstory trees. Using the canopy stratification, we developed a LiDAR pulse occlusion model in terms of point density and estimated that a density of about 170 pt/m² is required to segment understory trees found as deep as the third canopy layer as accurately as overstory trees. Lastly, we presented a distributed computing approach to scale up tree segmentation of an entire forest. The approach enabled locating individual tree and retrieving the point cloud segments representing tree crowns for large forested areas in a timely manner.

The presented work in this chapter provides methods for processing discrete return LiDAR point clouds of complex and large forested areas. Together with the advancements of the LiDAR sensor technology and platforms enabling acquisition of denser point clouds, it is expected that accurate and remote quantification of forest resources at the individual tree level both for overstory and understory becomes feasible in near future, which in turn facilitates accurate monitoring, management, and conservation activities of these resources.

## Acknowledgments

This work was supported by: 1) the Department of Forestry at the University of Kentucky and the McIntire-Stennis project KY009026 Accession 1001477, ii) the Kentucky Science and Engineering Foundation under the grant KSEF-3405-RDE-018, and iii) the University of Kentucky Center for Computational Sciences.

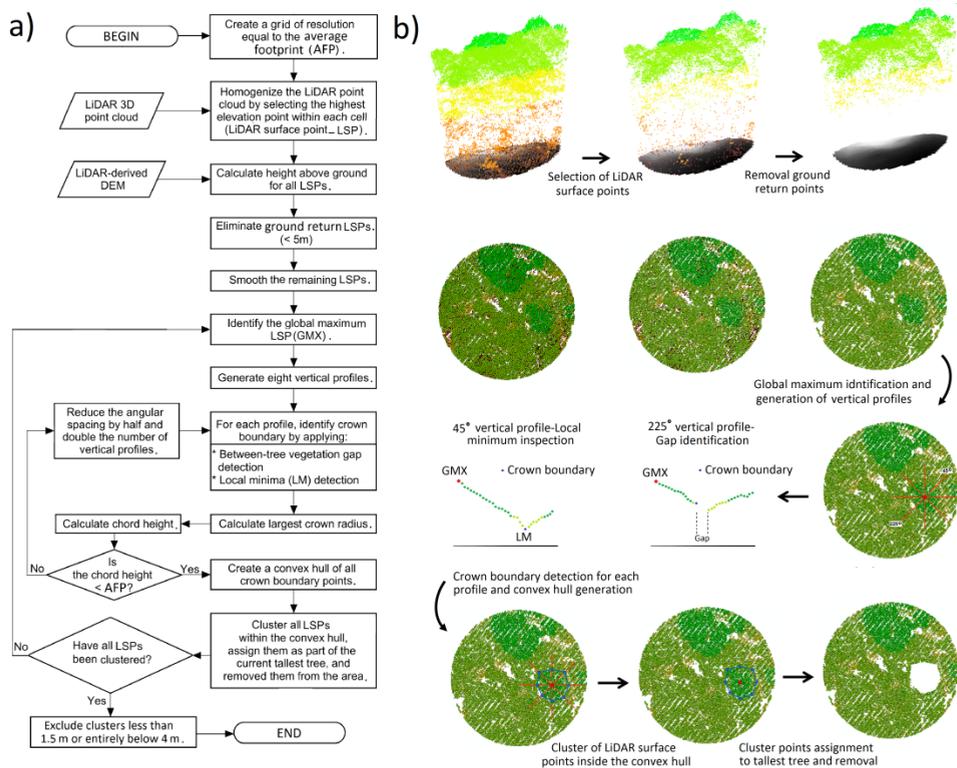

Figure 1. a) Flowchart of the tree segmentation method used to identify tree locations and delineate tree crowns.  b) illustration of the process of clustering the tallest tree crown.

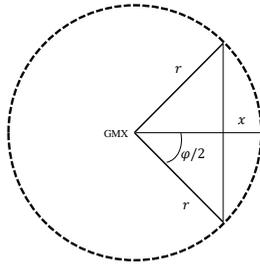

Figure 2. Diagram illustrating the calculation of the chord height ($x$) formed by two profiles of maximum crown radius ($r$) separated by the angular spacing ($\varphi$).

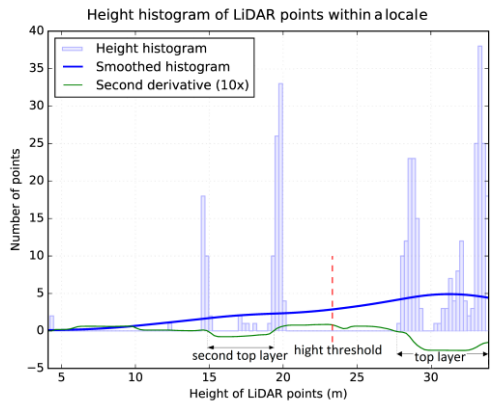

Figure 3. Height histogram of LiDAR points within a locale including over 100 points used for determining the height threshold for stratifying the top canopy layer in a cell location.

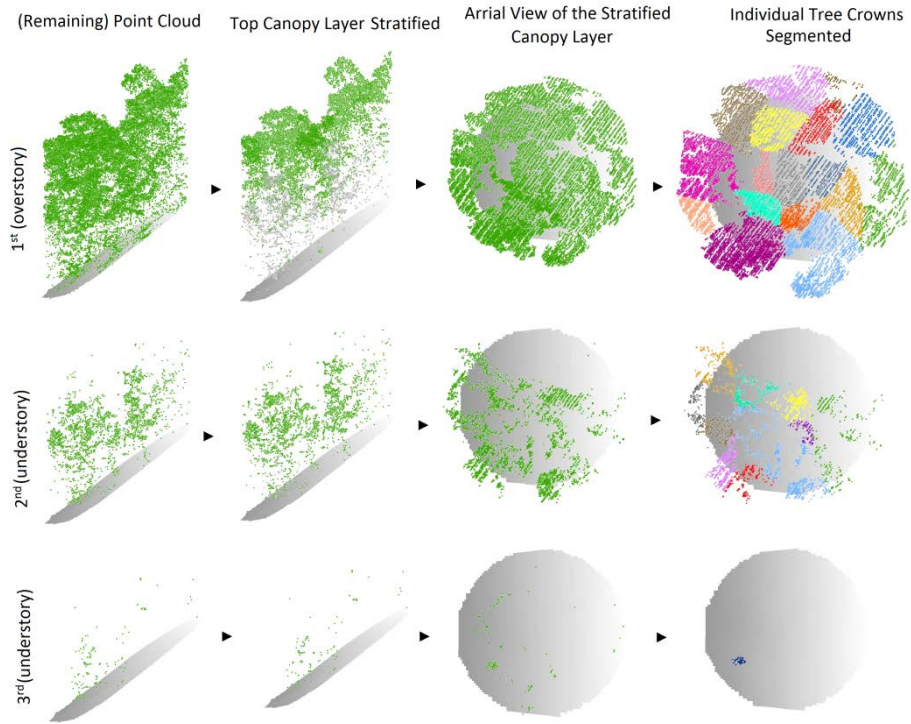

Figure 4. Illustration of the tree segmentation process in a multi-story stand by stratifying one canopy layer at a time, removing it from the point cloud, and segmenting crowns within it. A number of understory trees seem to be missed within the third canopy layer, which is likely due to the much lower point density compared to the first and second layers.

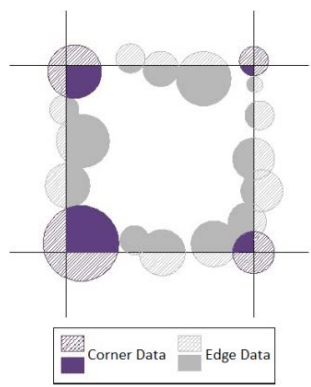

Figure 5. A schematic of a tile with the two types of boundary data. The solid-colored tree crown pieces inside the tile should be unified with the corresponding stripe-colored parts outside in a distributed computing environment.

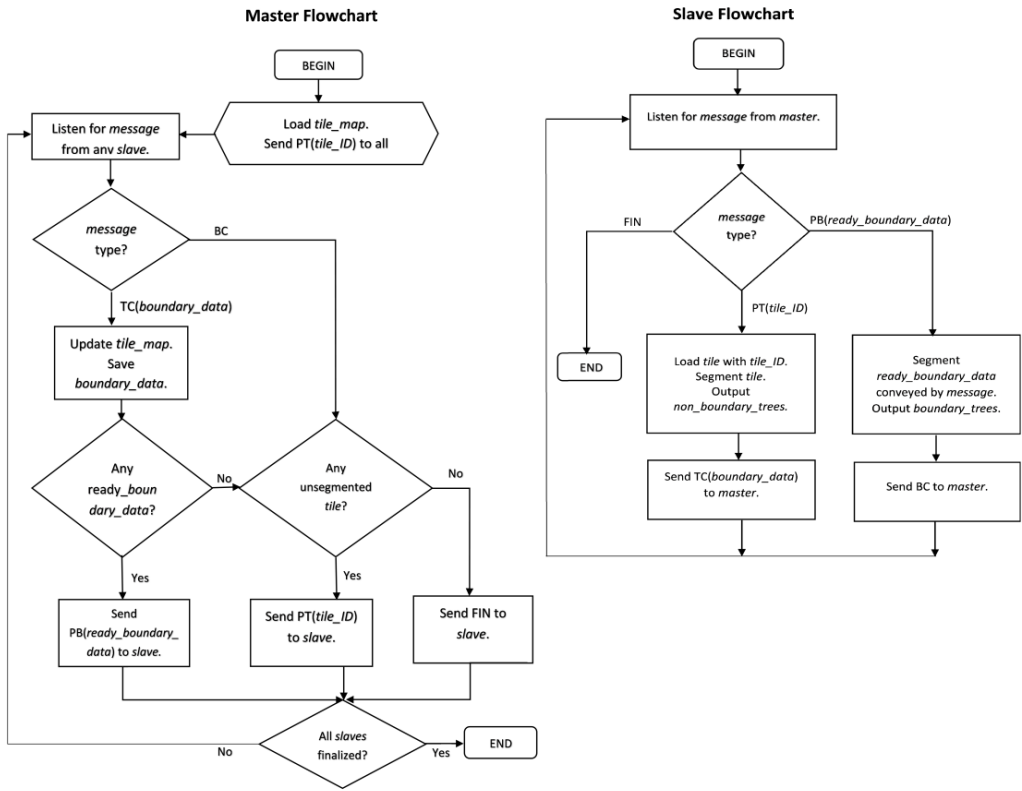

Figure 6. Flowcharts of the master and the slave. The master is responsible for maintaining the tile map globally and coordinating the slaves while a slave segments tiles and boundary data as directed by the master.

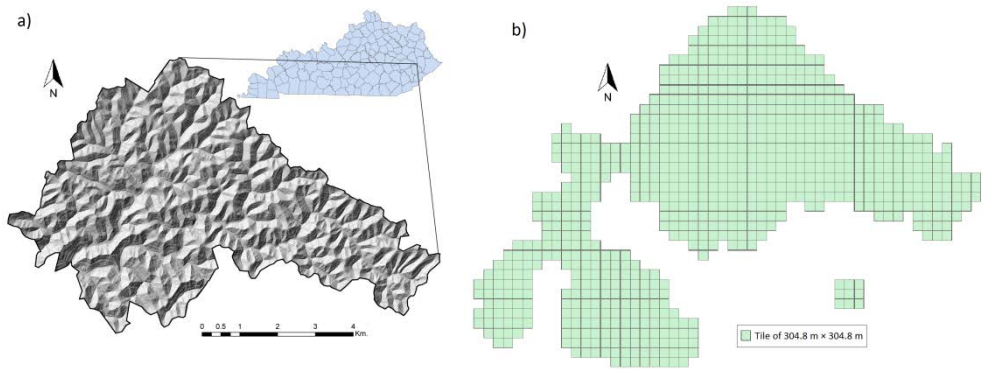

Figure 7. a) Terrain relief map of the University of Kentucky Robinson Forest and its general location within Kentucky, USA, b) LiDAR tile map of Robinson Forest consisting of 801 9.3-ha tiles.

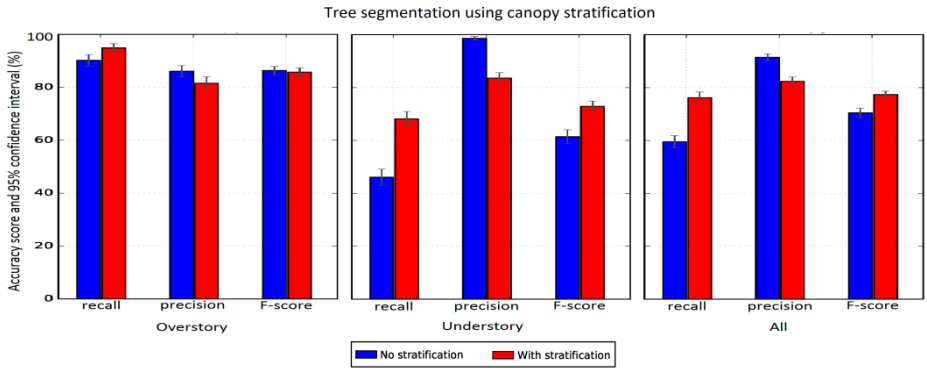

Figure 8. Average segmentation accuracies over 270 sample plots grouped by crown class.

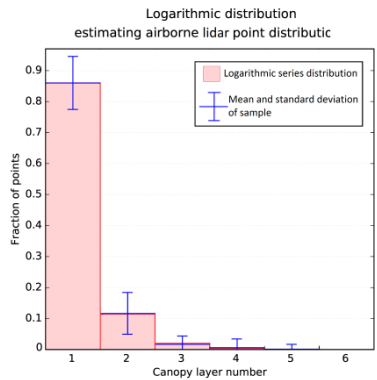

Figure 9. Logarithmic series distribution estimating observed fractions of LiDAR points recorded for different canopy layers. The distribution has a discrete domain supporting natural numbers.

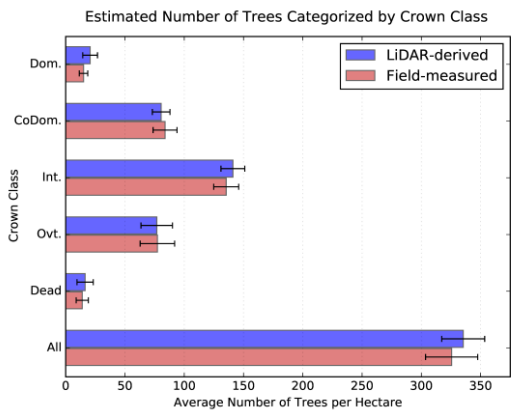

Figure 10. Estimated number of trees using LiDAR compared to field-collected along with the 95% confidence intervals.

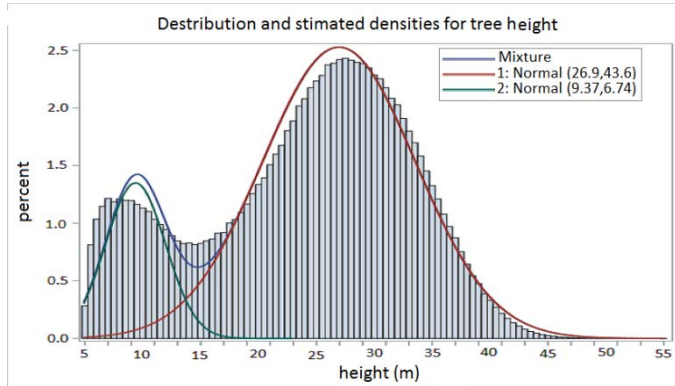

Figure 11. Height distribution of nearly two million trees segmented in Robinson Forest superimposed with estimated normal mixture model.

Table 1. Summary of plot level data collected from the 270 plots in Robinson Forest.

| Plot-Level Metric | | Min | Max | Avg. | Total | Percent of total |
|---|---|---|---|---|---|---|
| Slope | (%) | 0 | 93 | 50 | | |
| Aspect | º | 2 | 360 | 179 | | |
| Tree count | | 2 | 41 | 14.7 | 3,971 | |
| Dominant | | 0 | 3 | 0.5 | 130 | 3.3 |
| Co-dominant | | 0 | 10 | 3.5 | 954 | 24.0 |
| Intermediate | | 0 | 34 | 5.5 | 1,481 | 37.3 |
| Overtopped | | 0 | 19 | 4.3 | 1,152 | 29.0 |
| Dead | | 0 | 7 | 0.9 | 254 | 6.4 |
| Species count | | 1 | 12 | 6.0 | 43 | |
| Shannon diversity index | | 0.0 | 2.25 | 1.50 | | |
| Average tree Height | (m) | 13.9 | 28.8 | 19.5 | | |
| Standard deviation of tree heights | (m) | 1.2 | 12.4 | 5.5 | | |

Table 2. Summary statistics of the canopy layers stripped over the 270 sample plots.

| Canopy Layer | Plots[1] | Starting Height (m) | | Thickness (m) | | Point Density (pt/m²) | |
|---|---|---|---|---|---|---|---|
| | | Avg. | S.D. | Avg. | S.D. | Avg. | S.D. |
| 1 | 0.00% | 18.16 | 4.53 | 8.18 | 0.38 | 42.08 | 17.42 |
| 2 | 7.78% | 4.23 | 2.58 | 8.76 | 0.99 | 5.02 | 3.23 |
| 3 | 68.15% | 0.47 | 1.03 | 6.44 | 1.35 | 0.84 | 0.79 |
| 4 | 24.07% | 0.34 | 1.39 | 6.14 | 1.82 | 0.44 | 0.80 |
| Aggregate | 100.00% | 1.38 | 1.41 | 24.85 | 4.26 | 47.45 | 20.13 |

[1] Plots having as many number of canopy layers.

Table 3. Estimated number of trees categorized based on tree crown class.

| Crown Class | Fraction of existing to grand total detected | | Estimated number of existing trees | |
| --- | --- | --- | --- | --- |
| | mean | 95TCB[1] | entire forest | per ha |
| Dominant | 0.0785 | ±28.80% | 153,178 | 20.59 |
| Co-dominant | 0.3069 | ±7.84% | 599,106 | 80.50 |
| Intermediate | 0.5376 | ±8.18% | 1,049,446 | 141.32 |
| Overtopped | 0.2928 | ±10.94% | 571,522 | 76.80 |
| Dead | 0.0625 | ±18.63% | 121,917 | 16.38 |
| All | 1.2782 | ±5.71% | 2,495,170 | 335.30 |

[1] 95% T-Confidence Bounds (DF=269)